\documentclass[10pt,twocolumn,letterpaper]{article}

\usepackage[T1]{fontenc}

\usepackage[pagenumbers]{cvpr}

\usepackage[dvipsnames]{xcolor}

\definecolor{cvprblue}{rgb}{0.21,0.49,0.74}
\usepackage[breaklinks,colorlinks,citecolor=cvprblue]{hyperref}

\hypersetup{
  pdftitle={Multimodal Training with Knowledge Traction for Video Anomaly Detection},
  pdfauthor={Xiao Tan},
  pdfsubject={Weakly supervised video anomaly detection},
  pdfkeywords={weakly supervised video anomaly detection, multimodal learning, large language model, dictionary learning, cross-modal attention}
}

\usepackage{algorithm}
\usepackage{algorithmic}
\usepackage{booktabs}
\usepackage{amsmath}
\usepackage{amssymb}

\usepackage{multirow}
\usepackage{graphicx}
\usepackage{pifont}
\usepackage{tabularx}

\title{Multimodal Training with Knowledge Traction for Video Anomaly Detection}

\author{Xiao Tan\\
Xidian University, Xi'an, China\\
{\tt\small 21171213976@stu.xidian.edu.cn}}

\begin{document}
\maketitle
\begin{abstract}
The task of weakly supervised video anomaly detection is challenging. Existing approaches mainly model video frame features and determine whether anomalies are present based on their magnitudes. In this work, we propose a novel method that incorporates external knowledge into training to improve model performance, robustness, and generalization. We leverage textual prompts and a large language model (LLM) to establish prior information for the specific task, and employ dictionary learning to normalize and compress this information. Moreover, during experiments, we observed inconsistencies in the textual descriptions extracted from similar video frames by image caption models. To address this, we devise a method that leverages an LLM and auxiliary information to ensure consistency in the textual descriptions of similar frames within a video. Finally, we introduce a feature processing approach that incorporates prior knowledge and cross-modal attention mechanisms. This approach learns the relationships and characteristics among different modalities to construct more accurate and reliable feature representations, thereby improving the accuracy and stability of downstream tasks. Experimental results demonstrate that our method surpasses other state-of-the-art approaches.
\end{abstract}

\section{Introduction}
\label{sec:intro}

Anomaly detection in videos is a critical task with applications across many domains such as surveillance, healthcare, and autonomous driving\citep{a-liu2021, a-garg2020}. The goal is to identify irregular events that deviate from normal patterns in video data. However, annotated real-world video datasets containing anomalies are limited due to the prohibitive cost of exhaustive manual annotation. This has led to increasing interest in weakly supervised anomaly detection\citep{a-jiang2023weakly}, where models are trained on videos with only video-level labels indicating the presence of anomalies instead of frame-level or pixel-level annotations. 

\begin{figure}[t]
\centering
\includegraphics[width=\linewidth]{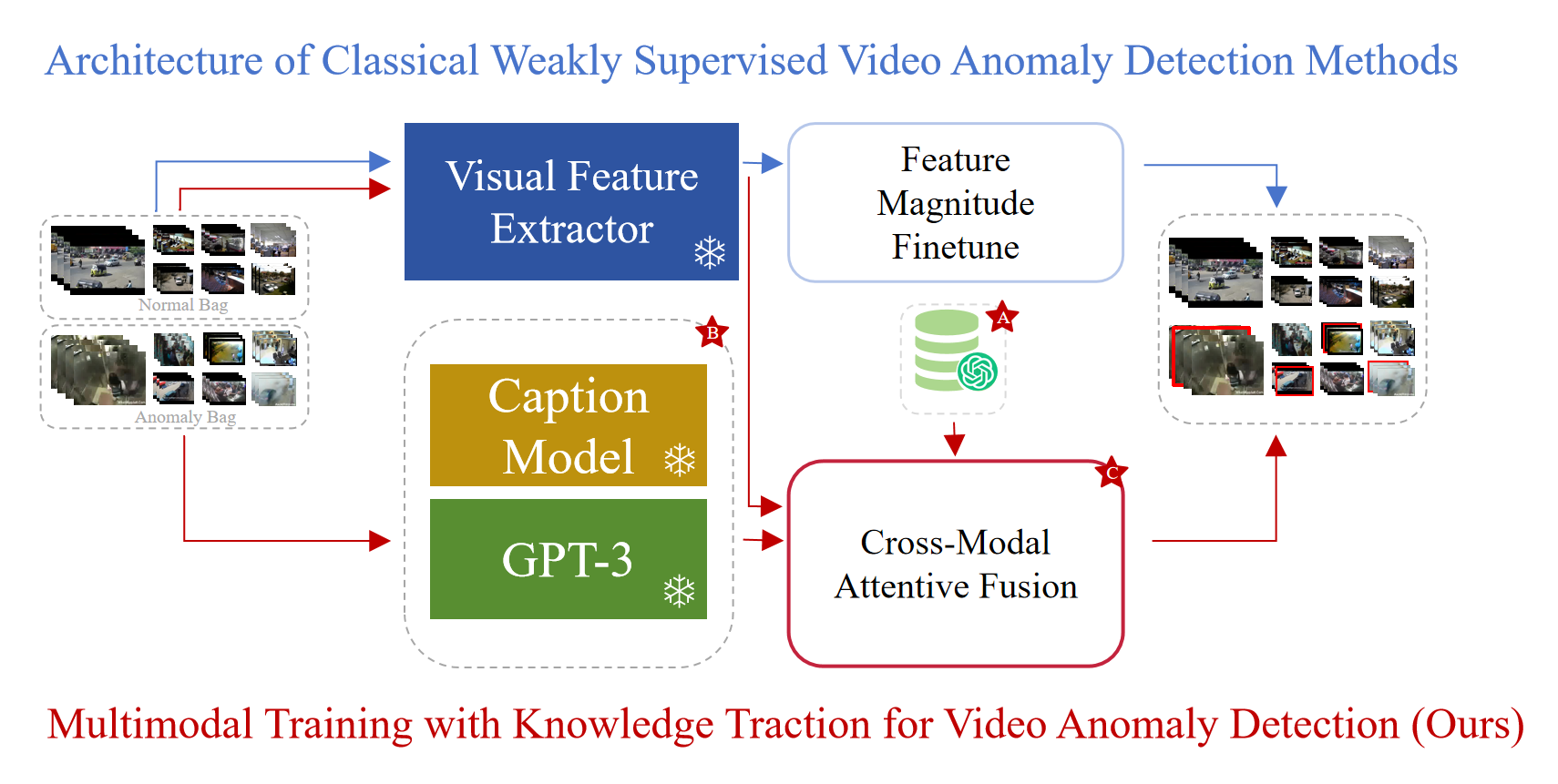}
\caption{The overall framework of the proposed method. The blue line is the data flow of the classical method and the red line is the data flow of our method. The three modules marked with stars are the novelties of our approach.}
\label{fig:arch}
\end{figure}

While weakly supervised methods reduce annotation costs, modeling normal activities and detecting anomalies from only video-level labels remains challenging. Existing approaches\citep{a-wu2022self,a-li2022self,a-tian2021weakly,a-sultani2018real} have focused on extracting discriminative features from individual frames and aggregating them to perform video-level classification. However, these methods do not explicitly model the temporal evolution of activities and often struggle to detect anomalies that persist over long durations. Recent work\citep{a-shin20203d,a-murugesan2020efficient} has incorporated temporal modeling using 3D ConvNets or recurrent neural networks (RNNs) over frame-level features. However, performance is still limited because these models lack sufficient supervision and tend to overfit to frequent normal activities while failing to identify infrequent anomalies effectively. Methods based solely on feature recognition or temporal information have reached a performance bottleneck. Therefore, as shown in Figure \ref{fig:arch}], we hypothesize that extracting richer information from videos and establishing prior knowledge for the targeted anomaly types, and then incorporating this knowledge as auxiliary supervision during training, would further improve the performance and generalization capability of the model.

In this work, we aim to improve weakly supervised anomaly detection by incorporating external knowledge from textual descriptions to provide richer supervision. Specifically, we leverage recent advances in large language models (LLMs) to generate textual prompts that describe normal activities and anomalies for a given dataset\citep{a-zhao2023survey}. These prompts prime the model with high-level semantic information about the appearance and evolution of events beyond what is contained in the video-level labels. 

However, directly incorporating raw prompt embeddings as additional input can be inefficient and redundant. Hence, we propose to first normalize the prompt embeddings using dictionary learning techniques\citep{a-tovsic2011dictionary} like Sparse Coding\citep{a-lee2006efficient} and Dense Associative Memory\citep{a-krotov2016dense}. This allows compressing the high-dimensional prompt embeddings into a compact dictionary of codewords that capture the essential semantic concepts. The codeword activations for a given prompt serve as its normalized representation. 

More formally, let $p_i \in \mathbb{R}^d$ be the embedding for prompt $i$ obtained from the LLM, where $d$ is the dimensionality. We learn a dictionary $D \in \mathbb{R}^{d \times K}$ with $K$ codewords that sparsely represents the prompt embeddings:

\begin{equation}
p_i \approx D a_i, \quad ||a_i||_0 \ll K
\end{equation}

where $a_i \in \mathbb{R}^K$ is the sparse code for prompt $i$ containing mostly zero activations, except for a few codewords relevant to the semantic content. We can compute the code activations efficiently using sparse coding algorithms like orthogonal matching pursuit. The code $a_i$ serves as the normalized prompt representation that will be provided as auxiliary input to the detection model.

Furthermore, during our experiments, we discovered a significant inconsistency issue in the scene text descriptions extracted from video frames, leading to ambiguity and distortion. This directly impacts the algorithm's ability to comprehend and recognize video content. To address this, we propose a novel consistency regularization method. For each video, we first obtain frame-level image captions using a pre-trained image captioning model such as BLIP\citep{a-li2022blip,a-li2023blip}. We then enforce consistency between frames with high similarity in the video by further using the LLM to generate the description from the original captions and extra supporting information generated by other models. Concretely, if $c_{i,1}, c_{i,2}, ... c_{i,T}$ are the frame-level captions for video $i$ and $d_{i,1}, d_{i,2}, ... d_{i,T}$ are extra supporting information gained by other models such as YOLO, we optimize: 

\begin{equation}
 c_{d_{i, 1}}, ..., c_{d_{i, T}} = \text{LLM}_\theta(c_{i,1}, ..., c_{i,T}, d_{i,1}, ..., d_{i,T})
\end{equation}

where $c_{d_{i, T}}$ is the textual description after LLM correction and $\theta$ denotes the LLM parameters. This regularization method corrects erroneous raw text descriptions with additional information.

For modeling, we propose a novel cross-modal architecture that leverages the normalized prompt embeddings. Our model consists of three branches: an RGB stream, an optical flow stream\citep{a-fortun2015optical}, and a text stream. The RGB and flow streams extract spatio-temporal features from frame sequences using pre-trained visual models\citep{a-carreira2017quo}. The text stream encodes the prompt representation $a_i$ and $c_{d_{i, T}}$ through an embedding layer followed by modeling with BART\citep{a-lewis2019bart}. 

We introduce cross-modal attention mechanisms that allow bi-directional interaction between the modalities. The text stream attends to relevant normal and anomalous patterns in the video streams using a text-visual attention. Conversely, the video streams focus on regions described in the text using a visual-text attention. This multi-stage attention exchange enables deeply fused integration of the complementary modalities. After that, the prior knowledge further optimizes the fused features by increasing or suppressing the expressiveness of different regions.

Finally, the fused features are passed through classification layers to predict frame-level anomalies. The cross-modal architecture is trained end-to-end using the video labels only. We empirically demonstrate significant gains over state-of-the-art approaches on multiple benchmark datasets. 

In summary, our key contributions are:

\begin{itemize}
    \item Leveraging LLMs to generate informative textual prompts as external knowledge and normalizing textual prompts via dictionary learning for efficient modeling
    \item Enforcing caption consistency using LLMs and extra supporting information
    \item Cross-modal neural architecture with bi-directional attentive fusion
    \item Demonstrating improved anomaly detection on two benchmark datasets
\end{itemize}

\section{Related Work}
\label{sec:relate}

This section reviews prior work on weakly supervised video anomaly detection and discusses relevant research on leveraging large language models and self-supervised dictionary learning for this task.

\subsection{Weakly Supervised Video Anomaly Detection}

Early work on weakly supervised video anomaly detection focused on handcrafted features and shallow models. For example, \citet{a-lu2013abnormal} extracted histogram of oriented gradients and optical flow features to train a multiple instance learning model. \citet{a-chen2020unsupervised} proposed an autoencoder trained on normal video frames to detect anomalies as frames with high reconstruction error. More recent methods have incorporated deep learning, training convolutional neural networks (CNNs) on video-level labels. \citet{a-zhong2019graph} introduced a graph convolutional network over sparse object detections. \citet{a-nguyen2019anomaly} proposed an autoregressive network with a composite memory module.

Other approaches have incorporated self-supervision by pretraining on proxy tasks before fine-tuning on anomaly detection. \citet{a-gong2019memorizing} pretrained a memory module on frame order verification. \citet{a-park2020learning} used a number of pretext tasks like rotation prediction. Contrastive predictive coding has also been effective for self-supervised pretraining \citep{a-liu2021weakly}. More sophisticated techniques include adversarial learning \citep{a-jin2021adversarial} and perturbation-based regularization \citep{a-morais2019learning}.

Overall, weakly supervised anomaly detection remains challenging, and existing methods typically suffer from high false positive rates. Incorporating external knowledge is a promising direction to improve performance.

\subsection{Large Language Models}

Large language models (LLMs) based on the Transformer architecture \citep{a-vaswani2017attention} have driven rapid progress in natural language processing. Models like BERT\citep{a-devlin2018bert}, GPT\citep{a-radford2018improving}, and T5\citep{a-raffel2020exploring} have achieved strong performance on a wide range of NLP tasks through self-supervised pretraining on large corpora.

More recently, even larger LLMs with over 100 billion parameters like GPT-3\citep{a-brown2020language} have shown the ability to perform few-shot and zero-shot learning. By formulating tasks as natural language prompts, GPT-3 can accomplish them without any gradient updates or fine-tuning. This suggests LLMs may serve as rich sources of world knowledge that can be tapped for downstream applications like video anomaly detection.

Methods to effectively incorporate LLMs include prompting \citep{a-liu2023pre} and adapter tuning \citep{a-houlsby2019parameter}. Prompting formulates the task as a text completion problem, while adapter tuning adds small task-specific modules to retain model capabilities. Both approaches allow leveraging the knowledge within LLMs without extensive fine-tuning.

\subsection{Self-Supervised Sparse Dictionary Learning}

Dictionary learning aims to find a representative set of patterns (a dictionary) to reconstruct inputs. Early dictionary learning methods were unsupervised \citep{a-aharon2006k}, but recent work has incorporated self-supervision.

\citet{a-shao2021ssdl} proposed a bootstrapping framework that alternates between dictionary construction and self-label assignment. The dictionary is initialized by clustering features, then used to assign pseudo-labels for self-supervision. This framework allows joint optimization of the dictionary and self-supervision.

Other methods like CURL \citep{a-laskin2020curl} and SwAV \citep{a-caron2020unsupervised} have incorporated dictionary learning objectives into contrastive self-supervised learning. The key idea is to map representations to a sparse code over a learnable dictionary. This provides a form of quantization that improves contrastive learning.

For video anomaly detection, dictionary learning can help extract common patterns from normal frames, while anomalies yield dense reconstructions. A self-supervised approach can construct a dictionary tailored to the normal data without manual labels. The dictionary can regularize models to focus on normal patterns and detect anomalies as outliers.

\section{Our Approach}
\label{sec:our}

\subsection{Generation of Prior Knowledge via LLM \label{sec:gen}}

\begin{figure}[htb]
\centering
\includegraphics[width=\linewidth]{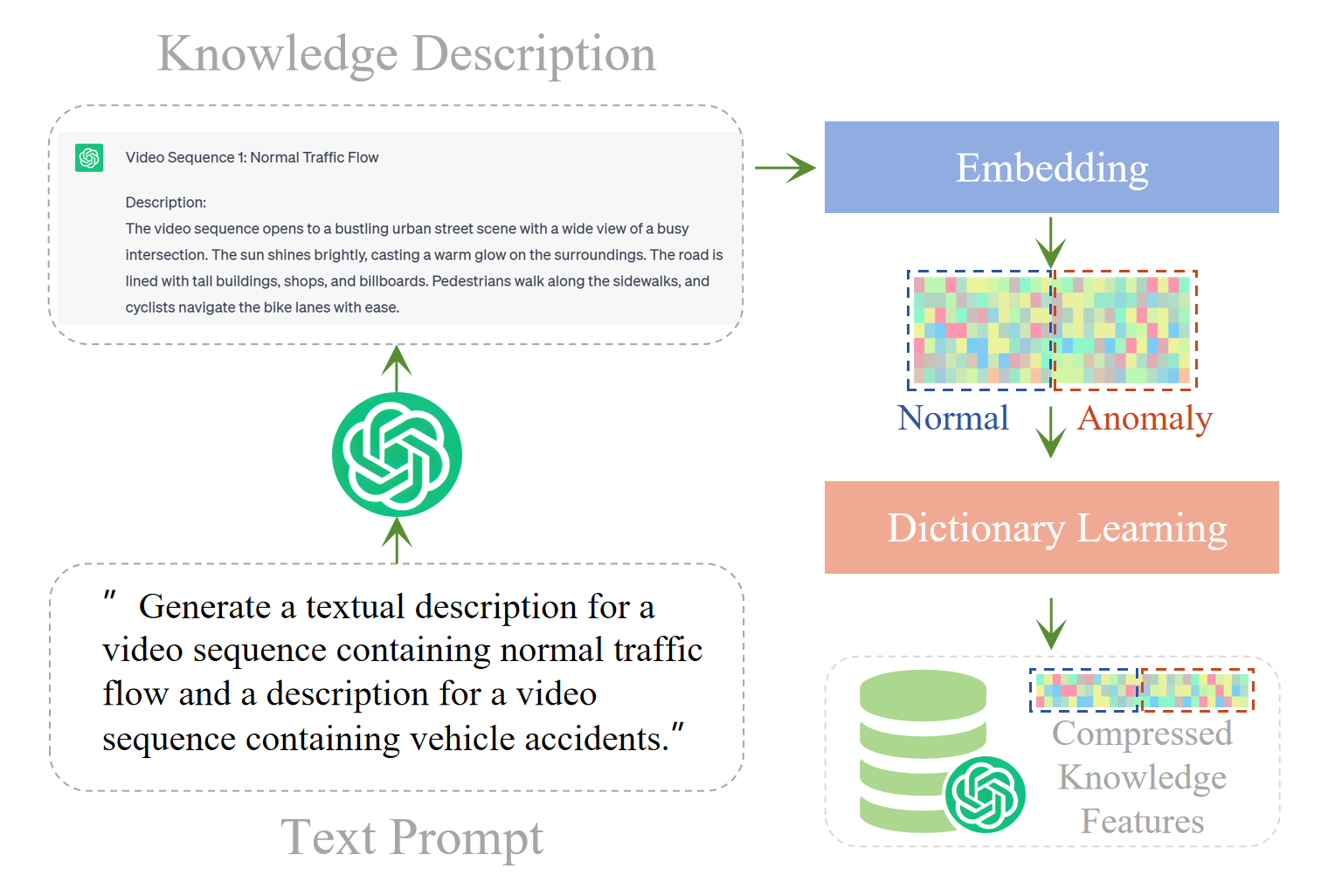}
\caption{Flowchart of the prior knowledge generation process. The LLM is inspired to generate knowledge in this domain by human-written domain-specific text prompts, which should include both normal-related and abnormal-related knowledge. The feature representation of the compressed knowledge is then obtained through text encoding and dictionary learning.}
\label{fig:kgen}
\end{figure}

An essential aspect of weakly supervised video anomaly detection is to incorporate external knowledge into the training process. As shown in Figure \ref{fig:kgen}], we propose a method for generating prior knowledge using an LLM to provide informative textual prompts related to different types of anomalous scenarios. We then utilize a pre-trained language model, such as BART\citep{a-lewis2019bart}, to embed the textual knowledge, followed by dictionary learning for feature purification and compression. The resulting compact and non-redundant feature representation serves as an external knowledge base for the subsequent training process.

\subsubsection{LLM-Based Knowledge Generation}

Given a video anomaly detection dataset, we first leverage GPT-3\citep{a-brown2020language} to generate textual knowledge with task-specific prompts, which provide high-level semantic descriptions conveying relevant information about both normal and anomalous activities in videos. 

For example, suppose we have the following query for a dataset targeting "vehicle accidents":

\begin{quote}
"Generate a textual description for a video sequence containing normal traffic flow and a description for a video sequence containing vehicle accidents."
\end{quote}

Then, GPT-3 generates two textual descriptions: one describing normal traffic flow and the other an anomalous sequence featuring vehicle accidents. These generated texts help exploit the knowledge encapsulated in the LLM and, in turn, improve anomaly detection. Formally, for the aforementioned query, it provides prompts $c_{normal}$ and $c_{anomaly}$.

\subsubsection{Knowledge Embedding via Dictionary Learning}

After obtaining the textual descriptions of the generated knowledge, BART\citep{a-lewis2019bart} is first used to embed the entire generated text and obtain $f_{normal}$ and $f_{anomaly}$. Since directly incorporating the raw embeddings can be computationally expensive and potentially redundant, we process these embeddings using dictionary learning. This compresses high-dimensional embeddings into a concise dictionary of codewords that effectively capture the core semantic content.

Let $e_i \in \mathbb{R}^d$ be the embedding for the textual prompt $i$ ($f_{normal}$ or $f_{anomaly}$) obtained from GPT-3 and BART, where $d$ is the dimensionality. Next, we train a dictionary $D \in \mathbb{R}^{d \times K}$ with $K$ codewords to sparsely represent the knowledge embeddings:

\begin{equation}
\min_{\mathbf{A}} \frac{1}{2} \| \mathbf{e_{i}} - \mathbf{D} \mathbf{A} \|_F^2 + \lambda \| \mathbf{A} \|_1
\end{equation}

where $\mathbf{A}$ is a coding matrix representing the sparse coding of each sample over the dictionary, $\| \cdot \|_F$ denotes the Frobenius norm, $\| \cdot \|_1$ denotes the L1 norm, and $\lambda$ is a regularization parameter controlling the sparsity of the encoding.

\subsection{Maintaining Consistency in Frame Descriptions \label{sec:maintain}}

\begin{figure}[htb]
\centering
\includegraphics[width=\linewidth]{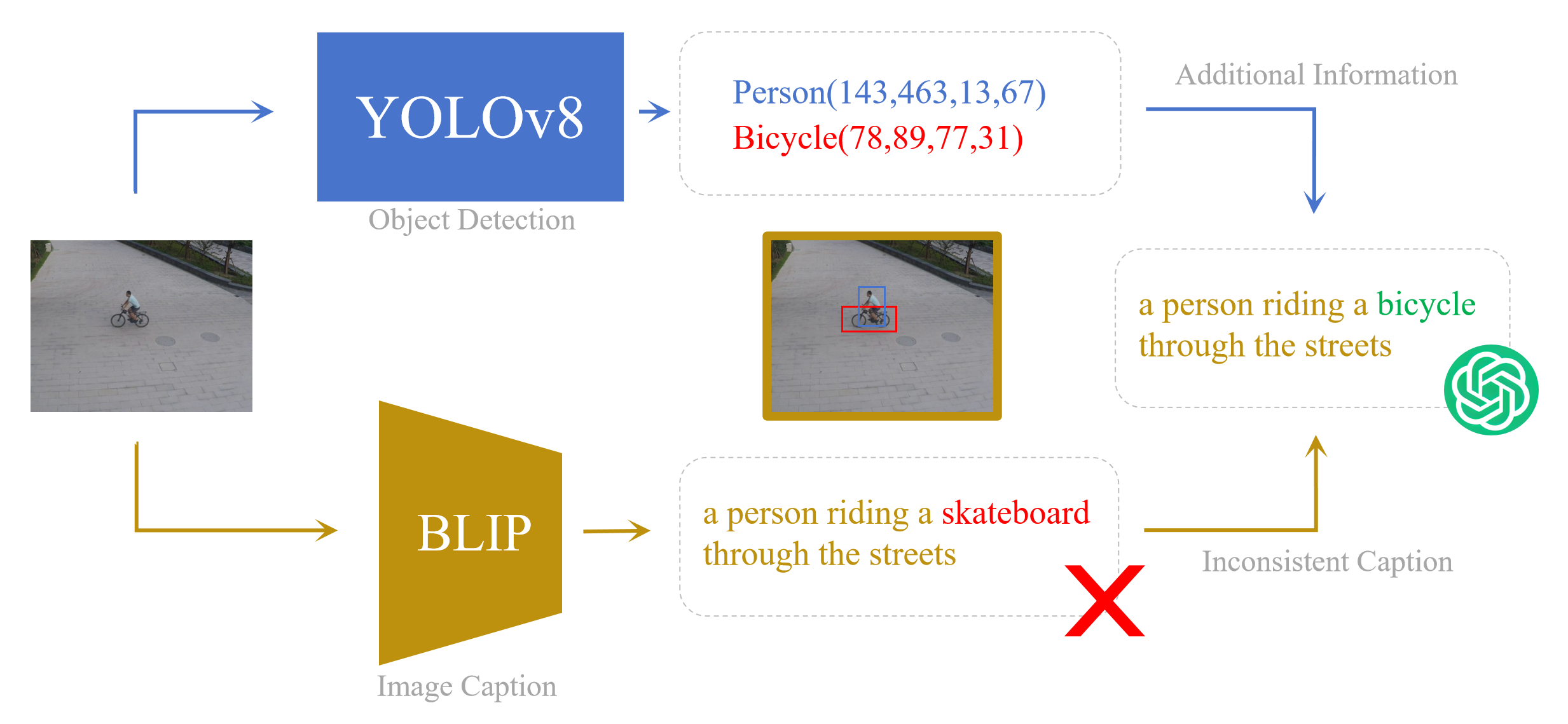}
\caption{With additional scene information provided by the object detection model, the LLM corrects incorrect frame descriptions generated by BLIP.}
\label{fig:maintain}
\end{figure}

As shown in Figure \ref{fig:maintain}], we address the issue of inconsistency in textual descriptions extracted from frames within a video by leveraging additional information from an object detection model. The inconsistency arises from the limitations of the image caption model in capturing fine-grained details and context in video frames. To mitigate this problem, we propose an approach that incorporates the output of the object detection model along with the frame descriptions and jointly processes them using an LLM. The LLM's powerful knowledge representation and logical reasoning capabilities enable us to correct and maintain consistency between the textual descriptions and the visual content of the frames.

Let \( F_i \) be a set of frames within a video, and \( T(F_i) \) be the corresponding textual descriptions generated by BLIP\citep{a-li2022blip}. We aim to incorporate the output of YOLOv8\citep{a-reis2023real}, denoted as \( D(F_i) \), which provides more accurate and detailed information about the objects present in the frames. We denote the corrected textual descriptions as \( \hat{T}(F_i) \), which are generated by the LLM as follows:

\begin{equation}
\label{eq:maintain}
\hat{T}(F_i) = \text{LLM}(T(F_i), D(F_i))
\end{equation}

In Equation (\ref{eq:maintain}), \( \text{LLM}(\cdot) \) denotes the LLM function. It takes the original textual descriptions \( T(F_i) \) and the object detection results \( D(F_i) \) as input and outputs the corrected descriptions \( \hat{T}(F_i) \).

\subsection{Cross-Modal Attention with Prior Knowledge}

\begin{figure*}[htb]
\centering
\includegraphics[width=\linewidth]{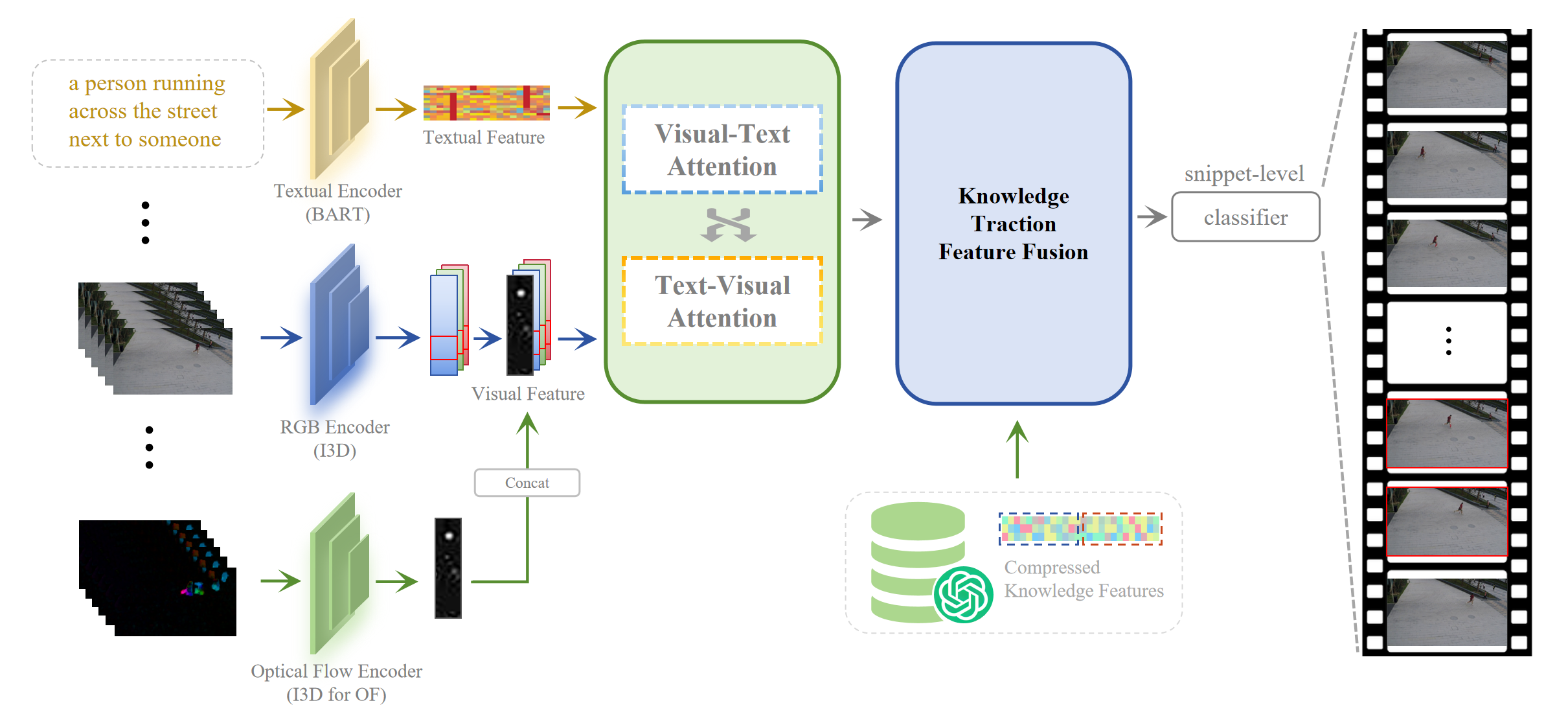}
\caption{Overview of the feature processing method. After the features of different modalities are encoded and concatenated, a bidirectional attention module performs deep feature extraction. The processed features are then optimized by the Knowledge Traction module and used for anomaly detection.}
\label{fig:cross}
\end{figure*}

\begin{figure}[htb]
	\begin{minipage}[t]{\linewidth}
		\centering
		\includegraphics[width=\linewidth]{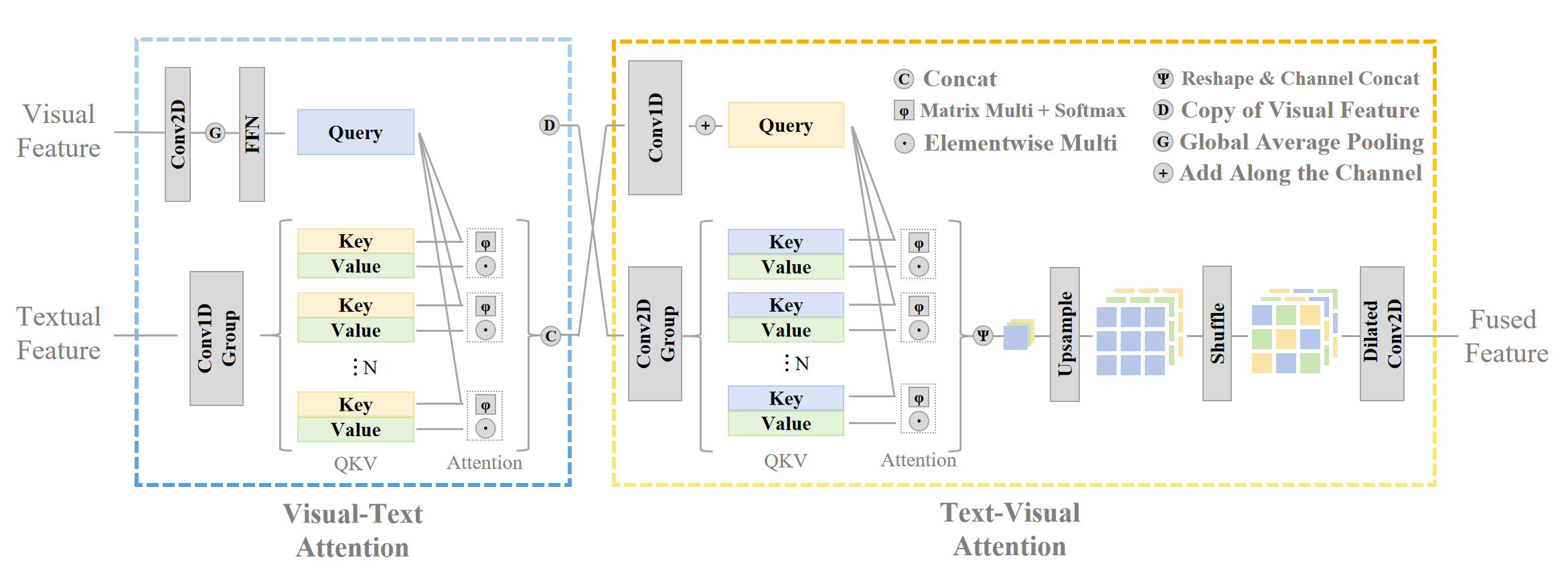}
		\subcaption*{(a) In the first stage, visual-text attention performs a coarse fusion of features. In the second stage, text-visual attention fine-tunes the fusion of features.}
		\label{fig:detail:attention}
	\end{minipage}
	\begin{minipage}[t]{\linewidth}
		\centering
		\includegraphics[width=\linewidth]{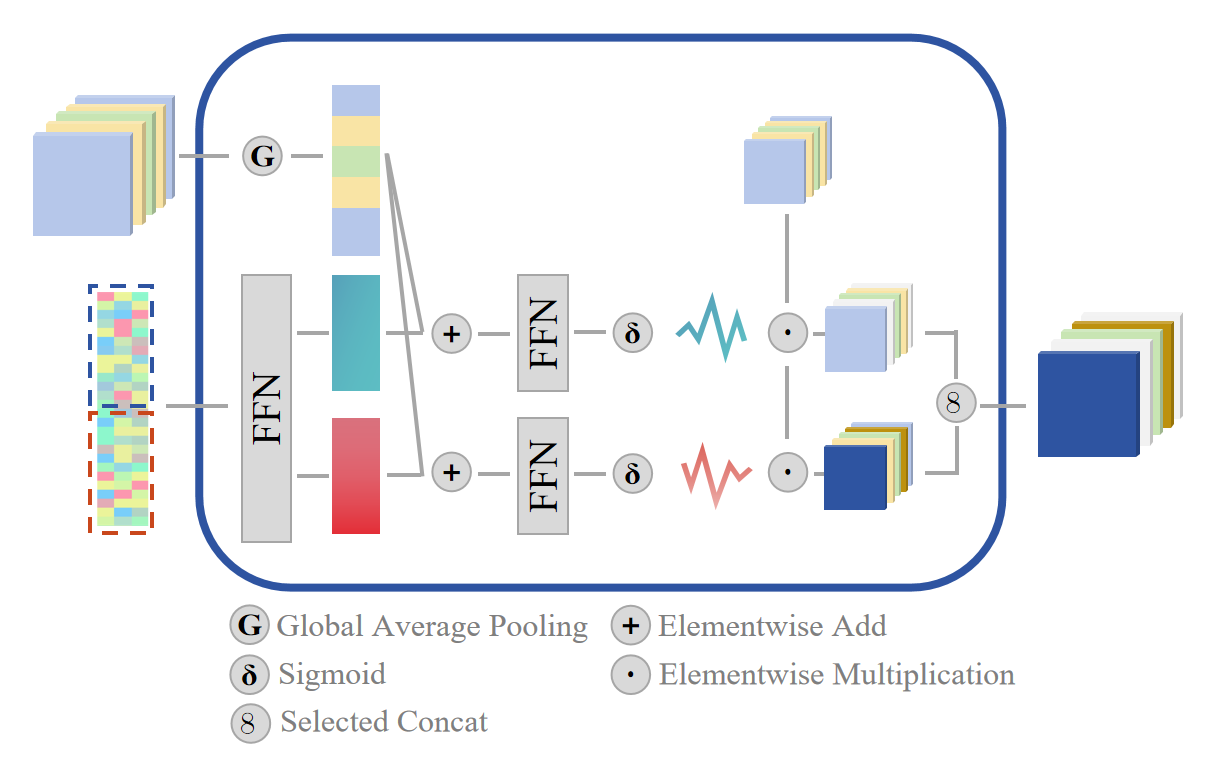}
		\subcaption*{(b) Amplify or suppress the different dimensions of the fused features based on prior knowledge, and further optimize the expressiveness of the features.}
		\label{fig:detail:fuse}
	\end{minipage}
\caption{Module implementation details. (a) The bidirectional attention module. (b) The Knowledge Traction module.}	
\label{fig:detail}
\end{figure}

This section details our cross-modal attention mechanism, which leverages prior knowledge obtained from external textual prompts. Its goal is to enable bidirectional interaction between the visual and textual modalities, allowing them to mutually enhance their representations based on the semantic information provided by the normalized textual prompts.

The architecture consists of three branches: the RGB stream, the optical flow stream, and the text stream. These branches process video frames, optical flow data, and the normalized textual prompts, respectively. Figure \ref{fig:cross} provides an overview of the proposed architecture.

The RGB stream and optical flow stream are responsible for extracting spatio-temporal features from the video frames. These streams utilize pre-trained visual models, such as I3D\citep{a-carreira2017quo}, to capture the visual patterns and motion information present in the video sequences. The text stream is encoded by BART\citep{a-lewis2019bart} and used to complement visual information and facilitate integration with prior knowledge in subsequent processes. Next, textual and visual features are fed into our Visual-Text \& Text-Visual Attention Module for deep feature fusion. Subsequently, the fused features are augmented or suppressed in different regions using prior knowledge, which guides further optimization of the feature representation.

\subsubsection{Visual-Text Attention}

As shown in Figure \ref{fig:detail}](a), the visual and textual features first enter the "Visual-Text Attention" module. The visual features are downsampled by a convolution operation and produce the query ($Q_{v}$) after global average pooling and mapping. In contrast, the text features are passed through N independent 1D convolutional layers for feature compression to reduce computational complexity, producing the key ($K_{t}^{i}$) and value ($V_{t}^{i}$). Independent convolutional layers also allow the model to learn from the text features from multiple perspectives, improving generalization. Finally, $K_{t}^{i}$, $V_{t}^{i}$, and $Q_{v}$ undergo the following attention operation:

\begin{align}
F_{t \mid v} &= \text{Concat}_{i\in \left [ 1, N \right ] }\left ( \frac{Q_{v} \times \left ( K_{t}^{i} \right ) ^{T}}{\sqrt{d_{k}} } \cdot V_{t}^{i} \right ) 
\end{align}

Unlike the classical Transformer\citep{a-vaswani2017attention}, we concatenate the output features of all branches and do not use an additional fully connected layer. This coarse attention operation preserves the features of the original modalities to a greater extent and helps the subsequent process perform deeper feature fusion.

\subsubsection{Text-Visual Attention}

The attention mechanism in the "Text-Visual Attention" module is similar to that in the first stage. The difference is that the features obtained in the first stage are used to generate $Q_{t}$, $$Q_{t} = \sum_{i= 1}^{N} \text{Conv1D}(F_{t \mid v}^{i})$$ and the initial visual features are used to generate $K_{v}$ and $V_{v}$. Since only a coarse feature fusion is performed in the first stage, the visual elements of its output features occupy only a small portion. Therefore, $Q_{t}$ can be considered as a representation of textual features in the second stage of the attention operation. Meanwhile, since it also fuses visual features, the second fusion stage is easier to converge and helps align visual and textual features.

In Figure \ref{fig:detail}](a), the merged feature $F_{v \mid t} =Attention(Q_{t},K_{v},V_{v})$ is reshaped and upsampled to restore the original dimensionality. Then, we design a shuffling method to shift different regions of different channels in the feature. We believe this enhances the model's ability to learn temporal and spatial information. Finally, a convolution operation produces $F_{fused}$.
\subsubsection{Knowledge Traction Feature Fusion}

After deep feature fusion, the resulting features are guided and optimized using prior knowledge, as shown in Figure \ref{fig:detail}](b). First, the feature vectors of normal and anomalous prior knowledge, $K_{n}$ and $K_{a}$, pass through a shared FFN, $FFN_{share}$, to extract common features. The resulting vectors are then summed with $GAP(F_{fused})$ and fed into $FNN_{n}$ and $FNN_{a}$ to extract discrepant features, producing $KV_{n}$ and $KV_{a}$. This process can be described by the following equation: $$KV_{n \parallel a } = FNN_{n \parallel a}(FNN_{share}(K_{n \parallel a}) + GAP(F_{fused}))$$ Then, a sigmoid function is applied to $KV_{n}$ and $KV_{a}$ to represent the degree of adaptation of $F_{fused}$ to $K_{n}$ and $K_{a}$. $F_{fused}$ is scaled according to this degree of adaptation to obtain $F_{n}^{'}$ guided by normal prior knowledge and $F_{a}^{'}$ guided by anomalous prior knowledge, respectively. Finally, the features with high activation in different time domains are selected from $F_{n}^{'}$ and $F_{a}^{'}$ and concatenated to obtain $F_{final}$.

\begin{align}
\label{eq:ktff:a}
F_{n \parallel a}^{'} &= \sigma (KV_{n \parallel a}) \times F_{fused} \\ 
\label{eq:ktff:b}
F_{final} &= \exists_{>\lambda }(F_{n}^{'}) \Join  \exists_{>\beta }(F_{a}^{'}) 
\end{align}

In Equations \ref{eq:ktff:a} and \ref{eq:ktff:b}, $\sigma$ represents the sigmoid function. $\exists_{>\chi}(\cdot)$ denotes selecting the part of the features whose intensity is greater than $\chi$. $\Join$ denotes the concatenation of the selected features while preserving their original order.

\subsection{Optimization Process}

Our optimization objective is adapted from RTFM\citep{a-tian2021weakly} with modifications for our method and consists of three parts: $\mathcal{L}_{cls}$, $\mathcal{L}_{sep}$, and $\mathcal{L}_{limit}$. $\mathcal{L}_{cls}$ denotes the snippet-level classification loss, which can be expressed as $$\mathcal{L}_{\text{cls}}(p = f_{\theta}(F_{final}), y) = \alpha BCE(\left \{ p  \right \}_{k}, y) + BCE(\hat{p}, y)$$, where $f_{\theta}$ denotes the classifier, $\{ \cdot \}_{k}$ denotes the $k$ highest magnitudes in the feature vector, $k=5$, and $\hat{p}$ denotes the remaining features.

$\mathcal{L}_{sep}$ is responsible for the separability of normal and abnormal samples. The distance between features is defined as $D_{k}(f^{+},f^{-}) = \sum_{k} \left \| \left \{ f^{+} \right \}_k  \right \|_{2} - \sum_{k} \left \| \left \{ f^{-} \right \}_k  \right \|_{2}$, and the constraint can be described as $L_{cons}(f^{+}, f^{-}) = \left | m - D(f^{+}, f^{-}) \right | + D_{\lambda \in (+,-)}(\{ f^{\lambda } \}_{k}, \{ f^{\lambda } \}_{k^{'}})$. The final loss is $$\mathcal{L}_{sep} = L_{cons}(F_{t \mid v}^{n}, F_{t \mid v}^{a}) + L_{cons}(F_{v \mid t}^{n}, F_{v \mid t}^{a})$$, where $\|\cdot\|_2$ denotes the L2 norm. $\{ \cdot \}_{k}$ and $\{ \cdot \}_{k^{'}}$ denote the $k$ largest and $k$ smallest magnitudes in the features, respectively.

$\mathcal{L}_{limit}$ constrains the degree to which prior knowledge bootstraps the features, preventing excessive bootstrapping from distorting the features. It is defined as $$\mathcal{L}_{limit} = \sum_{\chi \in (n,a)}^{} \left | \overline{F_{\chi}^{'}} - F_{\chi}^{'}  \right |$$ where $\overline{\cdot}$ denotes the mean of the feature. 

Finally, the overall objective is $\mathcal{L} = \lambda_1 \cdot  \mathcal{L}_{cls} + \lambda_2 \cdot  \mathcal{L}_{sep} + \lambda_3 \cdot  \mathcal{L}_{limit}$. We set $\lambda_1$, $\lambda_2$, and $\lambda_3$ to $1.0$, $0.01$, and $0.005$, respectively.

\section{Experiment}
\label{sec:exp}

\subsection{Experimental Setup}

\subsubsection{Datasets and Evaluation}

We conducted extensive experiments on the ShanghaiTech\citep{a-luo2017revisit} and UCF-Crime\citep{a-sultani2018real} datasets and compared many previous state-of-the-art methods\citep{a-wu2022self,a-li2022self,a-tian2021weakly,a-sultani2018real,a-zhong2019graph, a-zaheer2020claws,a-feng2021mist,a-lv2023unbiased, a-chen2023mgfn,a-wan2020weakly, a-wu2021learning,a-pu2022locality}. We use the area under the frame-level receiver operating characteristic curve (AUC) as the evaluation metric for all methods.

\subsubsection{Data Processing}

We use I3D\citep{a-carreira2017quo} to extract RGB and optical flow features from videos, and BLIP\citep{a-li2022blip} to generate frame descriptions, which are processed according to our approach. We generate textual prior knowledge for each of the two datasets using GPT-3\citep{a-brown2020language} and encode all text data using BART\citep{a-lewis2019bart}.

\subsubsection{Hyperparameter and Training}

We generate 5,000 normal and anomalous priors separately for each dataset and compress them into a 256-dimensional tensor using dictionary learning. All textual content is embedded as a 2048-dimensional tensor. We also set the number of convolutional groups in the attention module, $N$, to 8. Upsampling is performed using bilinear interpolation, and the channel dimensions are adjusted using a $1 \times 1$ convolution. For the last dilated convolution, we empirically set the dilation coefficient to 4. All experiments are conducted on an NVIDIA RTX 3090 GPU. Each model is trained for 15,000 iterations with a learning rate of 0.001. A more detailed experimental setup can be found in the Appendix.

\subsection{Comparison With State-of-the-Art Methods}

As shown in Tables \ref{tab:sht} and \ref{tab:ucf}, our method outperforms other state-of-the-art methods, including semi-supervised and weakly supervised works, on the ShanghaiTech and UCF-Crime datasets. 

\begin{figure}[htb]
\centering
\includegraphics[width=\linewidth]{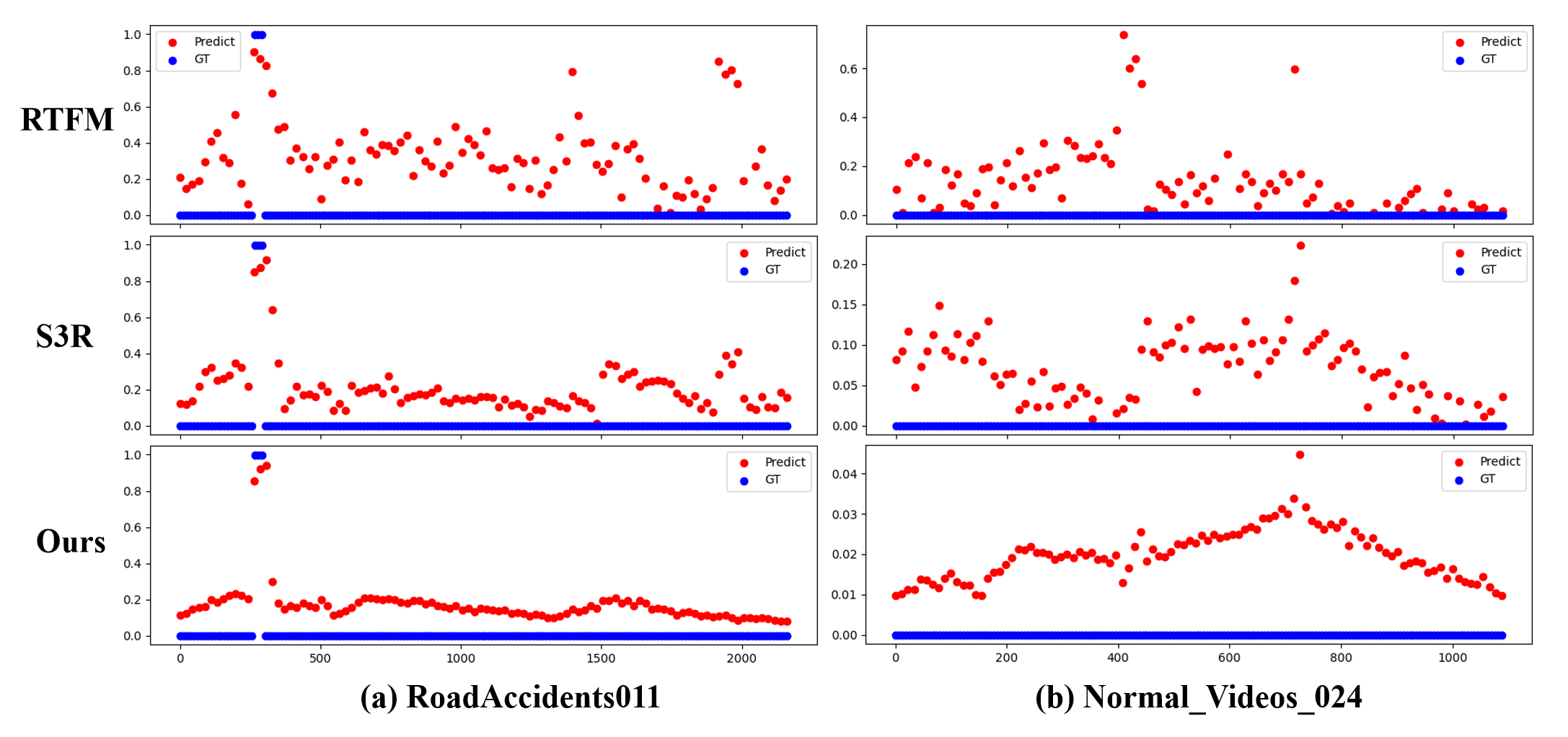}
\caption{Visualization of models' performance on UCF-Crime. The results show that our model has better accuracy and stability.}
\label{fig:compare}
\end{figure}

Specifically, our method improves performance by 7\% over AR-Net\citep{a-wan2020weakly}, which also uses optical flow information. Furthermore, our method achieves the best performance among all methods using I3D features. We believe this is because the visual and textual modalities complement each other during multimodal training, allowing the model to use the features more effectively. S3R\citep{a-wu2022self} also uses dictionary learning to clean up normal features to guide the model to better discriminate abnormal features. Unlike them, we use dictionary learning to compress the prior knowledge and let the model optimize the features for the scene under the guidance of the prior knowledge to improve performance. In Figure \ref{fig:compare}], we can see that our method performs better than other methods in terms of stability and accuracy, especially the results between adjacent normal frames are more continuous and stable. This may be because we maintain the consistency of text modalities across similar frames. More visualizations can be found in the Appendix.

Due to space limitations, please refer to the Appendix for more experimental details and visualizations.

\begin{table}[htb]
\centering
\resizebox{\columnwidth}{!}{%
\begin{tabular}{l|lll}
\hline
Supervision                         & Method   & Feature         & AUC (\%) \\ \hline
\multirow{3}{*}{Semi-supervised}    & Mem-AE   & -               & 71.2    \\
                                    & DLAN-AC  & -               & 74.7    \\
                                    & SVD-GAN  & -               & 78.42   \\ \hline
\multirow{11}{*}{Weakly-supervised} & MIL-Rank & $C3D_{rgb}$       & 86.3    \\
                                    & GCN      & $TSN_{rgb}$       & 84.44   \\
                                    & CLAWS    & $C3D_{rgb}$       & 89.67   \\
                                    & AR-Net   & $I3D_{rgb+flow}$  & 91.24   \\
                                    & MIST     & $I3D_{rgb}$       & 94.83   \\
                                    & CRFD     & $I3D_{rgb}$       & 97.48   \\
                                    & RTFM     & $I3D_{rgb}$       & 97.21   \\
                                    & MSL      & $VideoSwin_{rgb}$ & 97.32   \\
                                    & S3R      & $I3D_{rgb}$       & 97.48   \\
                                    & UML      & $XCLIP_{rgb}$    & 96.78   \\ \cline{2-4} 
                                    & \textbf{Ours}     & $I3D_{rgb+flow}$  & \textbf{98.17}   \\ \hline
\end{tabular}%
}
\caption{Performance comparison of state-of-the-art methods on the ShanghaiTech dataset.}
\label{tab:sht}
\end{table}

\begin{table}[htb]
\centering
\resizebox{\columnwidth}{!}{%
\begin{tabular}{l|lll}
\hline
Supervision                         & Method   & Feature         & AUC (\%) \\ \hline
\multirow{3}{*}{Semi-supervised}    & Conv-AE  & -               & 50.6    \\
                                    & S-SVDD   & -               & 58.5    \\
                                    & GODS     & $TCN_{rgb}$       & 70.46   \\ \hline
\multirow{11}{*}{Weakly-supervised} & MIL-Rank & $C3D_{rgb}$       & 75.41   \\
                                    & GCN      & $TSN_{rgb}$       & 82.12   \\
                                    & MIST     & $I3D_{rgb}$       & 82.3    \\
                                    & RTFM     & $I3D_{rgb}$       & 84.3    \\
                                    & DDL      & $I3D_{rgb}$       & 85.12   \\
                                    & MSL      & $I3D_{rgb}$       & 85.3    \\
                                    & S3R      & $I3D_{rgb}$       & 85.99   \\
                                    & MGFN     & $VideoSwin_{rgb}$ & 86.67   \\
                                    & MGFN     & $I3D_{rgb}$       & 86.98   \\
                                    & UML      & $XCLIP_{rgb}$    & 86.75   \\ \cline{2-4} 
                                    & \textbf{Ours}     & $I3D_{rgb+flow}$  & \textbf{87.11}   \\ \hline
\end{tabular}%
}
\caption{Performance comparison of state-of-the-art methods on the UCF-Crime dataset.}
\label{tab:ucf}
\end{table}

\subsection{Ablation Studies}

We also conducted a series of ablation experiments to study the effects of different modules and parameter settings on our method.

Table \ref{tab:module} shows the performance impact of the different modules in our method. When only Text-Visual Attention is used, performance improves significantly. However, when Knowledge Traction is introduced without Visual-Text Attention, performance deteriorates. We believe this is because the output of Text-Visual Attention is largely a visual feature and does not fit well with the textual prior knowledge. In contrast, after introducing Visual-Text Attention, the text modality participates more in feature fusion through attention, so the final fused features are better aligned with the textual prior knowledge and Knowledge Traction plays a leading role.

We also investigate the effect of the number of prior knowledge items generated by the LLM. As shown in Table \ref{tab:know}, we conduct experiments with different numbers of prior knowledge items and also investigate model performance when the amounts of normal and abnormal knowledge are unbalanced. The results show that when the amount of prior knowledge is small, performance degrades, possibly because the Knowledge Traction module cannot optimize the features correctly. On the other hand, if the amount of prior knowledge is too large, performance decreases slightly. We found that when too much prior knowledge is generated, many descriptions are repetitive, which introduces noise and affects model performance. In addition, the results show that unbalanced amounts of normal and abnormal knowledge negatively affect the model. Clearly, the Knowledge Traction module is also sensitive to this imbalance. Based on these results, we set the number of generated prior knowledge items to 5,000.

Table \ref{tab:nn} presents our exploration of the number of convolutions in the attention module. Experiments show that our method achieves the best performance when $N = 8$. Too small a number can lead to insufficient feature fusion, while too large a number can lead to overfitting and reduced generalization.

\begin{table}[htb]
\centering
\resizebox{\columnwidth}{!}{%
\begin{tabular}{llll|ll}
\hline
Baseline & VTA & TVA & KT & ShanghaiTech   & UCF-Crime      \\ \hline
\ding{52}        & \ding{56}   & \ding{56}   & \ding{56}  & 92.36          & 83.31          \\
\ding{52}        & \ding{56}   & \ding{52}   & \ding{56}  & 97.71          & 86.33          \\
\ding{52}        & \ding{56}   & \ding{52}   & \ding{52}  & 93.04          & 82.69          \\
\ding{52}        & \ding{52}   & \ding{52}   & \ding{56}  & 97.66          & 86.74          \\
\ding{52}        & \ding{52}   & \ding{52}   & \ding{52}  & \textbf{98.17} & \textbf{87.11} \\ \hline
\end{tabular}%
}
\caption{Impact of different modules of our approach on performance. VTA stands for Visual-Text Attention, TVA for Text-Visual Attention, and KT for Knowledge Traction.}
\label{tab:module}
\end{table}

\begin{table}[htb]
\centering
\resizebox{\columnwidth}{!}{%
\begin{tabular}{ll|ll}
\hline
Normal Count & Anomaly Count & ShanghaiTech   & UCF-Crime      \\ \hline
2500         & 2500          & 97.41          & 85.97          \\
2500         & 5000          & 97.83          & 86.59          \\
5000         & 2500          & 97.28          & 86.41          \\
5000         & 5000          & \textbf{98.17} & \textbf{87.11} \\
10000        & 10000         & 97.95          & 86.93          \\ \hline
\end{tabular}%
}
\caption{Influence of the number of prior knowledge items on performance.}
\label{tab:know}
\end{table}

\begin{table}[htb]
\centering
\resizebox{\columnwidth}{!}{%
\begin{tabular}{l|ll}
\hline
Number of Convolutions & ShanghaiTech   & UCF-Crime      \\ \hline
1                      & 96.23          & 84.89          \\
2                      & 97.19          & 85.76          \\
4                      & 97.64          & 86.88          \\
8                      & \textbf{98.17} & \textbf{87.11} \\
16                     & 97.78          & 86.72          \\ \hline
\end{tabular}%
}
\caption{Effect of the number of convolutions in the attention module on performance.}
\label{tab:nn}
\end{table}

\section{Conclusion}
\label{sec:conc}

In this work, we proposed a novel approach to weakly supervised video anomaly detection that incorporates external knowledge generated by a large language model and exploits cross-modal attention mechanisms. In addition, we addressed the issue of inconsistency in textual descriptions and proposed a cross-modal architecture with bidirectional attentive fusion that effectively integrates visual and textual information. Experimental results on benchmark datasets show that our approach outperforms other state-of-the-art methods. Overall, our approach provides a promising direction for advancing weakly supervised video anomaly detection using external knowledge and cross-modal modeling.

{
    \small
    \bibliographystyle{ieeenat_fullname}
    \bibliography{main}
}

\clearpage
\appendix
\section*{Appendix}
\renewcommand{\thefigure}{A\arabic{figure}}
\renewcommand{\thetable}{A\arabic{table}}
\renewcommand{\thealgorithm}{A\arabic{algorithm}}
\setcounter{figure}{0}
\setcounter{table}{0}
\setcounter{algorithm}{0}

\section{Algorithm Description}

This section provides a pseudocode description of the proposed method.

\begin{algorithm}[htb]
\caption{Algorithm for prior knowledge generation}
\label{alg:kgen}
\textbf{Input $Prompt_{n}$,$Prompt_{a}$}: Human-written prompt text\\
\textbf{Parameter $LLM$}: Large language model\\
\textbf{Parameter $Encoder$}: Pre-trained models for text embedding\\
\textbf{Parameter $N_{n}$}: Number of normal prior knowledge items to generate\\
\textbf{Parameter $N_{a}$}: Number of anomalous prior knowledge items to generate\\
\textbf{Output}: Features of prior knowledge \\
\begin{algorithmic}[1] 
\STATE Let $i=0$,$j=0$
\STATE Let $normalTexts$, $anomalyTexts$
\WHILE{i $ < N_{n}$}
\STATE $normalTexts$.add($LLM(Prompt_{n})$)
\STATE $i += 1$
\ENDWHILE
\WHILE{j $ < N_{a}$}
\STATE $anomalyTexts$.add($LLM(Prompt_{a})$)
\STATE $j += 1$
\ENDWHILE

\STATE \textbf{return} $Encoder(normalTexts)$ 
\STATE \quad \quad \quad $Encoder(anomalyTexts)$
\end{algorithmic}
\end{algorithm}

Algorithm \ref{alg:kgen} describes how to generate and embed prior knowledge in text form. The algorithm takes domain-specific textual prompts and generates descriptions of what is normal and what is abnormal. Because random hints are added, the LLM output has a low repetition rate, so this algorithm can generate a large amount of prior knowledge.

\begin{algorithm}[htb]
\caption{Algorithm for maintaining consistency in frame descriptions}
\label{alg:maintain}
\textbf{Input $Prompt$}: Human-written prompt text for fixing errors in description\\
\textbf{Input $Frames$}: Pending video frames\\
\textbf{Parameter $LLM$}: Large language model\\
\textbf{Parameter $BLIP$}: Pre-trained model for image captioning\\
\textbf{Parameter $YOLO$}: Pre-trained object detection model \\
\textbf{Output}: Corrected video frame description text \\
\begin{algorithmic}[1] 
\STATE Let $i=0$,$Corrected$
\WHILE{i $ < len(Frames)$}
\STATE description $= BLIP(Frames[i])$
\STATE extra $= YOLO(Frames[i])$
\STATE $Corrected$
\STATE \quad.add($LLM([Prompt, description, extra])$)
\STATE $i += 1$
\ENDWHILE

\STATE \textbf{return} $Corrected$
\end{algorithmic}
\end{algorithm}

Algorithm \ref{alg:maintain} describes how to maintain the consistency and correctness of the textual descriptions of video frames. The effectiveness of our method relies on the assumption that information produced by object detection models is more accurate. Therefore, we can use it and an LLM to correct the descriptions generated by BLIP.

\begin{algorithm}[htb]
\caption{Algorithm for cross-modal attention with prior knowledge}
\label{alg:cross}
\textbf{Input $F_{t}$}: Text features\\
\textbf{Input $F_{v}$}: Visual features\\
\textbf{Parameter $F_{k}^{n}$,$F_{k}^{a}$}: Prior knowledge features\\
\textbf{Parameter $N$}: In the attention module, the number of convolutions in the convolution group\\
\textbf{Output}: Fused features\\
\begin{algorithmic}[1] 
\STATE Let $i=0$,$j=0$
\STATE $Q_{v} = FFN(GAP(Conv2D(F_{v})))$
\WHILE{i $ < N$}
\STATE $K_{t} = FFN_{k}(Conv1D[i](F_{t}))$
\STATE $V_{t} = FFN_{v}(Conv1D[i](F_{t}))$
\STATE $F_{t \mid v} = Concat(F_{t \mid v}, Attention(Q_{v}, K_{t}, V_{t}))$
\STATE $i += 1$
\ENDWHILE
\STATE $Q_{t} = add(Conv1D(F_{t \mid v}), dim=1)$
\WHILE{j $ < N$}
\STATE $K_{v} = FFN_{k}(GAP(Conv2D[j](F_{v})))$
\STATE $V_{v} = FFN_{v}(GAP(Conv2D[j](F_{v})))$
\STATE $F_{v \mid t} = Concat(F_{v \mid t}, $
\STATE \quad \quad \quad $Reshape(Attention(Q_{t}, K_{v}, V_{v})))$
\STATE $j += 1$
\ENDWHILE
\STATE $F_{att} = DConv2D(Shuffle(Upsample(F_{v \mid t})))$
\STATE $rate_{n} = Sig(FFN_{n}(FFN(F_{k}^{n}) + GAP(F_{att})))$
\STATE $rate_{a} = Sig(FFN_{a}(FFN(F_{k}^{a}) + GAP(F_{att})))$
\STATE $F_{fused} = Concat(Select(F_{att}*rate_{n}), $
\STATE \quad \quad \quad \quad \quad \quad \quad \quad $Select(F_{att}*rate_{a}))$

\STATE \textbf{return} $F_{fused}$
\end{algorithmic}
\end{algorithm}

Algorithm \ref{alg:cross} illustrates the cross-modal attention mechanism and how prior knowledge optimizes the features.

\section{Examples of Prior Knowledge}

This section presents the generated priors for different scenarios. Due to space limitations, only a subset of the generated results is shown.

Table \ref{tab:sht-know} shows prior knowledge generated by GPT-3 for the ShanghaiTech dataset. The scenes in this dataset are relatively simple, consisting of campus pedestrian streets. In these scenes, normal behavior is usually walking, while other behaviors such as biking or playing are abnormal. Given these dataset characteristics, we carefully design the prompt text to encourage the LLM to generate descriptions that satisfy the requirements of prior knowledge as much as possible. We also emphasize randomness in the prompt text, which maximizes the variety of generated content.

Table \ref{tab:ucf-know} shows examples generated for several anomaly types in the UCF-Crime dataset. Thanks to GPT-3's large knowledge base and reasoning ability, and under the deliberate guidance of our prompt text, rich descriptions are generated for specific anomaly types. During prior knowledge generation, we noticed that some generated texts overlap to varying degrees. However, we did not filter these patterns because descriptions from different perspectives may increase the robustness of the prior knowledge, and dictionary learning may also mitigate the impact of excessive redundancy on performance.

For the distribution of generated prior knowledge, we adopt a balanced strategy. For ShanghaiTech, since the scenes and anomalies are relatively simple, we do not subdivide anomaly types and let the LLM generate an equal number of normal and abnormal samples. For UCF-Crime, the scenes are more complex and anomaly types are more diverse, so abnormal samples account for one third of all samples, with an equal number of samples for each anomaly type. This makes the characteristics of each anomaly type more distinct and helps the model detect anomalies.

\begin{table*}[htb]
\centering
\begin{tabularx}{\textwidth}{lXXX}
\hline
Dataset & Prompt Text & Normal Description                                    & Anomalous Description                                              \\ \hline
\multirow{6}{*}{ShanghaiTech} &
  \multirow{6}{4.5cm}{I need to detect the abnormality of a certain street in the campus. On this street, only normal walking is normal. Other situations, such as passing by bicycle, passing by fast running, etc., are abnormal. Please help me to randomly generate text descriptions of normal and abnormal situations. Note that it is randomly generated.} &
  "People leisurely walking down the street" &
  "A person riding a bicycle through the pedestrian area." \\
        &             & "Students chatting while strolling along the sidewalk."   & "Someone roller-skating at a high speed down the sidewalk."          \\
        &             & "A group of friends walking together at a moderate pace." & "A jogger sprinting through the street segment."                     \\
        &             & "Someone carrying a backpack and walking calmly."         & "Individuals playing a game of frisbee in the middle of the street." \\
        &             & "Students walking with their hands in their pockets."     & "Someone skateboarding recklessly down the sidewalk. "               \\
        &             & "A student wearing headphones, walking at a steady pace." & "A student running with a backpack on their back."                   \\ \hline
\end{tabularx}
\caption{Examples of prior knowledge generated by the LLM on the ShanghaiTech dataset.}
\label{tab:sht-know}
\end{table*}

\section{Detailed Experimental Setup}

The main text describes part of our experimental setup; due to space limitations, some details are provided here.

For data processing, we extract video frames at 16 FPS and resize them to 224$\times$224. For frames with a different aspect ratio, we use zero padding. We process these frames using I3D and embed frames from the same video into a tensor of shape $(T, N, C)$, where $T$ is the number of frames, $N=10$ indicates 10-crop data augmentation, and $C=2048$ is the feature dimension. The same scheme is used to extract optical flow features after preprocessing the video frames.

For text data, after obtaining frame descriptions with our method, we use BART to embed each frame description and obtain a tensor of shape $(T, C)$, where $T$ is the number of video frames and $C=2048$ is the feature dimension. Since video durations differ, we quantize $T$ for both visual and text features to 32. Our textual prior knowledge is also embedded using BART, resulting in a tensor of shape $(N, C)$, where $N$ is the number of prior knowledge items generated by the LLM and $C=2048$ is the embedding dimension. We then apply dictionary learning to obtain a tensor of shape $(N^{'}, C)$, where $N^{'}=256$.

For model configuration, we compress both visual and textual features by a factor of $\frac{1}{8}$, reducing the dimension from 2048 to 256 to lower computational cost. We also use dropout with a rate of 0.7. In our attention module, for the coarse-grained "Visual-Text Attention", we set the FFN output dimension to 256. In contrast, for the fine-grained "Text-Visual Attention", we set the FFN output dimension to 512. We set the number of convolutions in the convolution group to 8 based on our experiments. Finally, upsampling uses bilinear interpolation, and the last dilated convolution has a dilation coefficient of 4.

For training, we use the Adam optimizer with a learning rate of 0.001 and weight decay of 0.005, without learning rate decay. We set the batch size to 32 and train for 15,000 iterations. The final model is the one with the highest performance on the validation set. The configuration of our method is summarized in Table \ref{tab:train-cfg}.

\begin{table}[htb]
\centering
\resizebox{\columnwidth}{!}{%
\begin{tabular}{l|ll}
\hline
Type                               & Item                      & Value    \\ \hline
\multirow{5}{*}{Visual processing} & FPS                       & 16       \\
                                   & Resize                    & 224      \\
                                   & Padding                   & 0        \\
                                   & Augmentation              & 10-crop  \\
                                   & Feature dimension         & 2048     \\ \hline
\multirow{3}{*}{Text processing}   & Feature dimension         & 2048     \\
                                   & Quantize dimension        & 32       \\
                                   & Dictionary dimension      & 256      \\ \hline
\multirow{7}{*}{Model}             & Feature compression ratio & $\frac{1}{8}$      \\
                                   & Dropout ratio             & 0.7      \\
                                   & Coarse dimension          & 256      \\
                                   & Fine-tuned dimension      & 512      \\
                                   & Number of convolutions    & 8        \\
                                   & Upsampling                & bilinear \\
                                   & Dilated coefficient       & 4        \\ \hline
\multirow{5}{*}{Training}          & Optimizer                 & Adam     \\
                                   & Learning rate             & 0.001    \\
                                   & Weight decay              & 0.005    \\
                                   & Batch size                & 32       \\
                                   & Iteration                 & 15000    \\ \hline
\end{tabular}%
}
\caption{Detailed hyperparameter configuration of our method.}
\label{tab:train-cfg}
\end{table}

\section{Visualization of Experimental Results}

\begin{figure*}[htb]
\centering
\includegraphics[width=\linewidth]{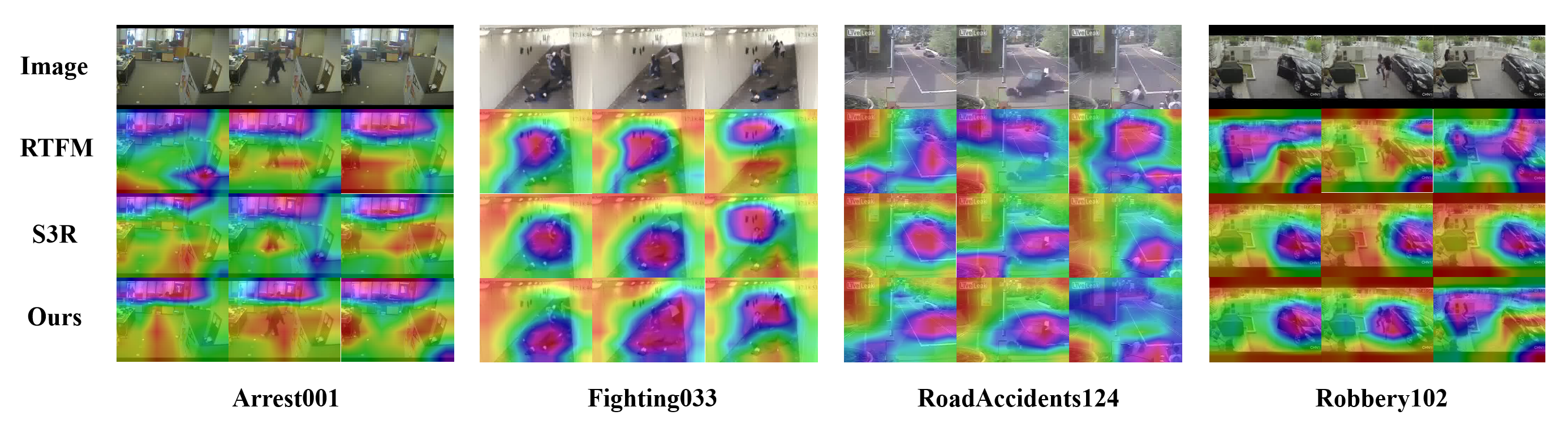}
\caption{Visualization of anomalous activation maps for some samples in the UCF-Crime dataset.}
\label{fig:app-cam}
\end{figure*}

This section presents additional visualization results and analyzes them.

\begin{figure}[htb]
\centering
\includegraphics[width=\linewidth]{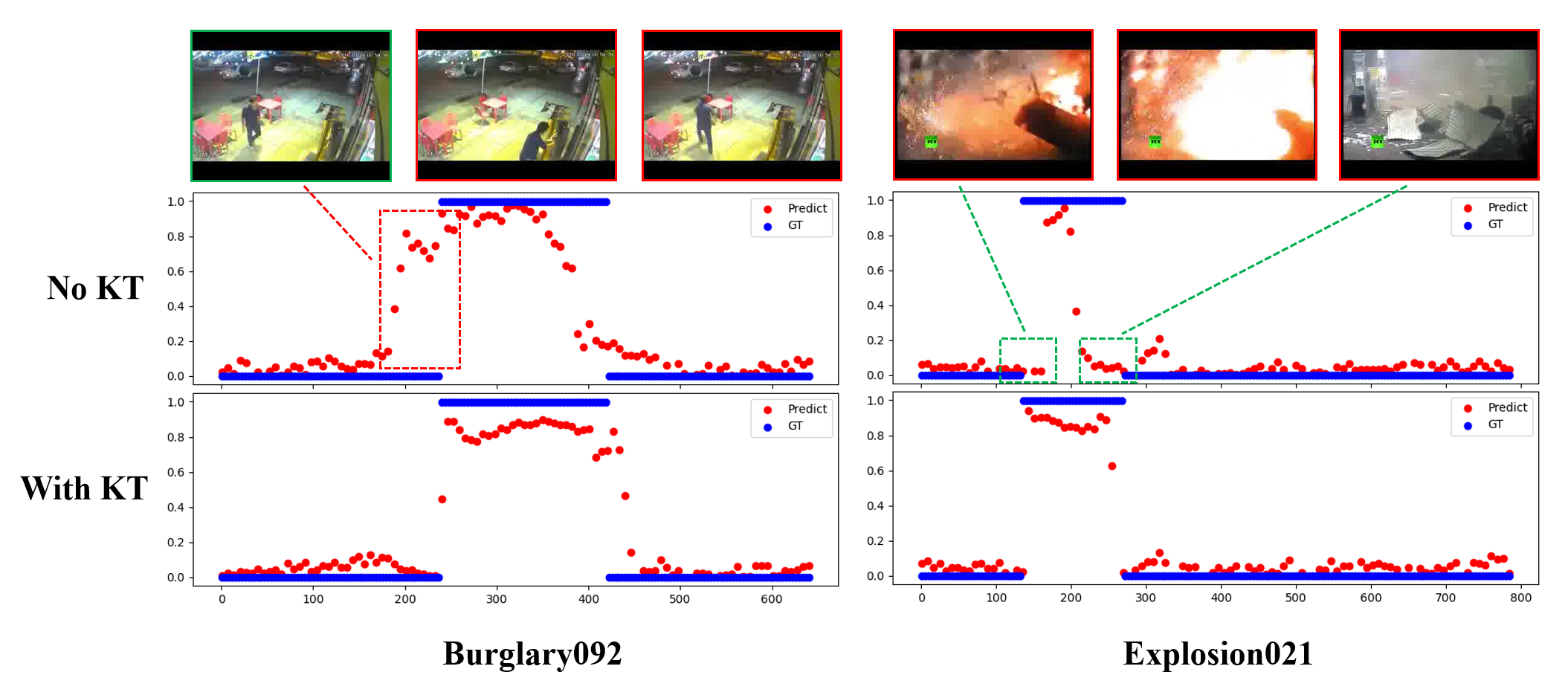}
\caption{Visualization of the impact of the Knowledge Traction (KT) module on model performance. For abnormal videos, it improves model robustness.}
\label{fig:app-know-ab}
\end{figure}

\begin{figure}[htb]
\centering
\includegraphics[width=\linewidth]{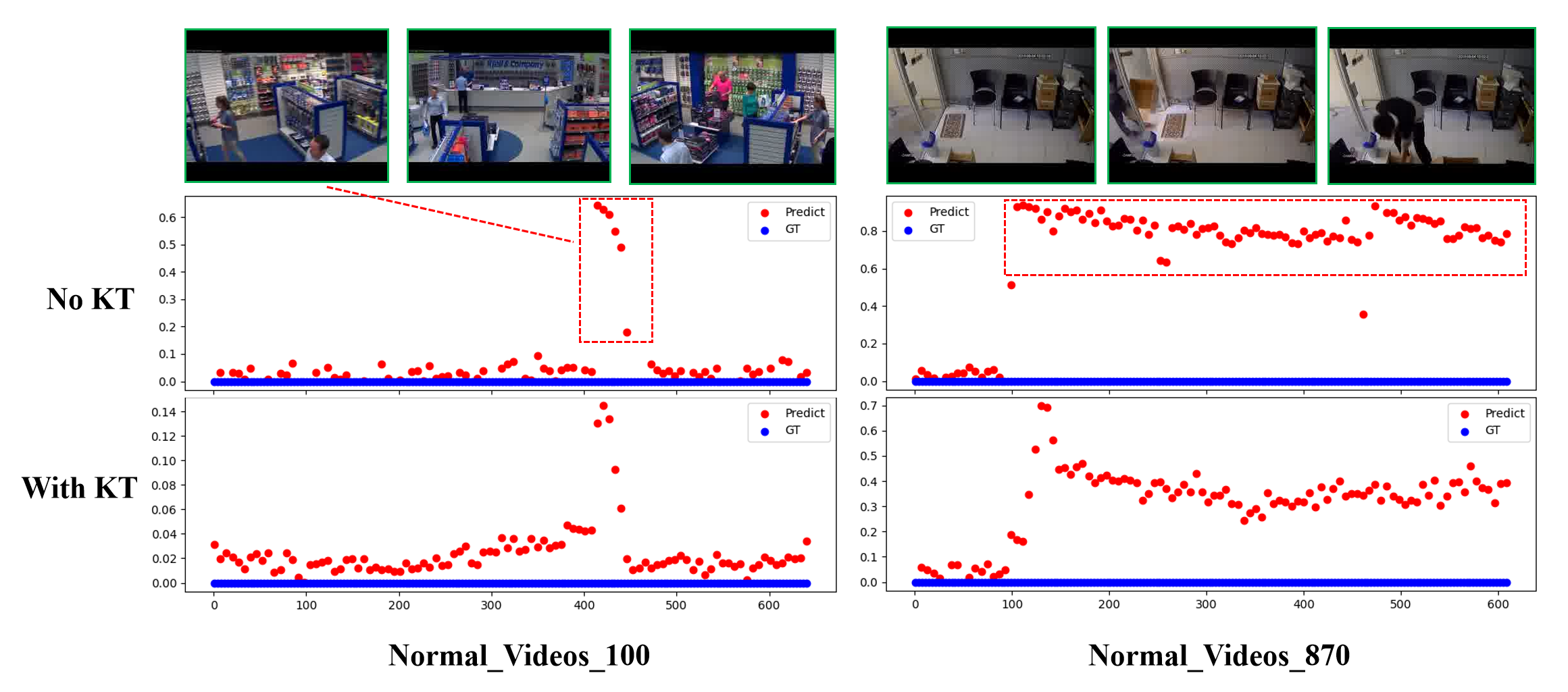}
\caption{Visualization of the impact of the Knowledge Traction (KT) module on model performance. For normal videos, it reduces the model's sensitivity to scene changes.}
\label{fig:app-know-no}
\end{figure}

Figures \ref{fig:app-know-ab} and \ref{fig:app-know-no} show additional visualizations that demonstrate the superiority of our Knowledge Traction module. In the "Burglary092" sample, the Knowledge Traction module corrects wrongly detected frames by suppressing the activation intensity of normal frames through prior knowledge. In the "Explosion021" example, some abnormal frames are clearly recognized as normal. A possible reason is that the color of the explosion scene is too uniform, leading to inaccurate recognition. After adding the Knowledge Traction module, this issue is effectively alleviated because, with prior knowledge, the text modality guides the model to understand the scene information in the video more comprehensively during training.

In contrast, some normal videos, such as "Normal\_Videos\_100" and "Normal\_Videos\_870", contain fast scene changes or abnormal-looking scenes that cause the model to fail. After introducing the Knowledge Traction module, the model still makes errors in these cases, but the magnitude of the detected anomalies is reduced.

\begin{figure}[htb]
\centering
\includegraphics[width=\linewidth]{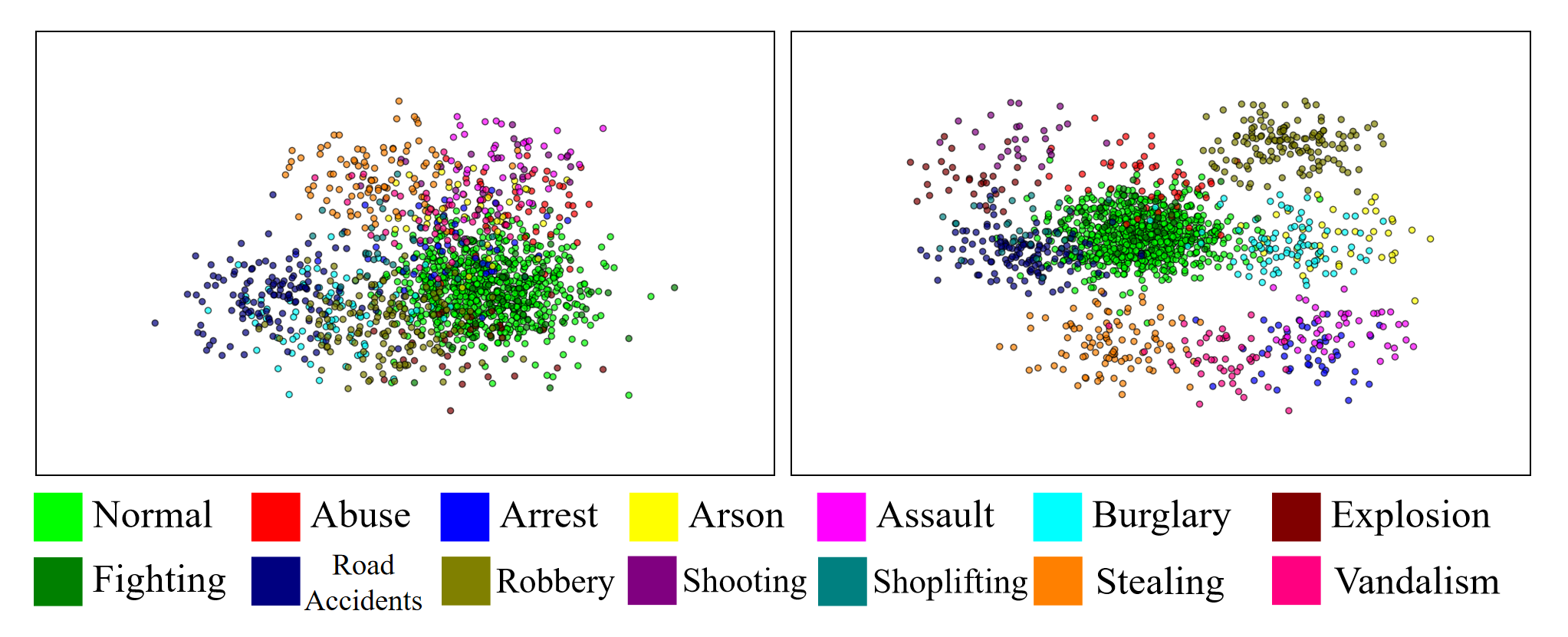}
\caption{2D visualization of features from the UCF-Crime dataset. Comparison between our method and the baseline MIL method.}
\label{fig:app-scatter}
\end{figure}

In Figure \ref{fig:app-scatter}, we visualize the features generated by the model on UCF-Crime by reducing their dimensionality. These features are labeled with 14 categories, comprising one normal label and 13 abnormal labels. The visualization shows that our model's features exhibit improved discrimination among categories. This improvement can be attributed to the effective fusion of visual and textual information within our features. Consequently, during training, our model gradually develops a deeper understanding of the distinctions between different anomaly types. This observation further highlights the benefits of multimodal fusion, as it enables the model to extract superior features from the data. Additionally, we visualize the activation map of a sample, as shown in Figure \ref{fig:app-cam}. The results reveal that our method concentrates more precisely on the region where the anomaly occurs. At the same time, this concentration is accompanied by less divergence than alternative approaches. Notably, our model successfully activates certain objects within the video frame even when they occupy a relatively small area. This emphasizes the effectiveness of our model in capturing meaningful anomalies.


\begin{table*}[p]
\centering
\begin{tabularx}{\textwidth}{lXXX}
\hline
Scenes &
  Prompt Text &
  Normal Description &
  Anomalous Description \\ \hline
\multirow{3}{*}{Steal} &
  \multirow{3}{4.5cm}{I need to detect anomalies related to theft. The detection scenes are usually the interior scenes of shops or houses, which are rich and varied. Please help me randomly generate text descriptions of normal and abnormal scenarios.} &
  "A well-lit retail store with shelves neatly arranged, customers browsing products, and employees assisting them." &
  "A dimly lit retail store with broken shelves, items scattered on the floor, and a person acting suspiciously and concealing items in their bag." \\
 &
   &
  "A cozy living room with comfortable furniture, family members watching TV, and soft music playing in the background." &
  "A disheveled living room with overturned furniture, shattered glass, and signs of forced entry through a broken window." \\
 &
   &
  "A quiet bookstore with customers peacefully reading, classical music playing, and a barista serving coffee at the in-store cafe." &
  "A chaotic bookstore with books strewn all over the floor, a fire alarm blaring, and an individual stuffing books into a bag hurriedly." \\
  \hline
\multirow{3}{*}{Explosion} &
  \multirow{3}{4.5cm}{I need to detect anomalies about explosions. The detection scenarios are rich and diverse. The cause of the explosion could be various reasons. Please help me randomly generate text descriptions of normal and abnormal scenarios.} &
  "A bustling parking lot on a sunny day, cars coming and going as people run errands." &
  "A deafening explosion shatters the quietness of the parking lot, sending debris and smoke into the air." \\
 &
   &
  "A quiet residential street with children playing, neighbors chatting, and birds chirping." &
  "Chaos erupts on the residential street as a building collapses due to a gas leak, causing panic and confusion." \\
 &
   &
  "A serene gas station with vehicles filling up, attended by workers in their uniforms." &
  "An intense fire breaks out at the gas station, flames engulfing fuel pumps and sending plumes of black smoke into the sky." \\
  \hline
\multirow{3}{*}{RoadAccidents} &
  \multirow{3}{4.5cm}{I need to detect anomalies about road accidents. The detection scene is generally outdoors, which is rich and varied. The types of accidents are equally diverse. Please help me randomly generate text descriptions of normal and abnormal scenarios.} &
  "Vehicles smoothly flow along the avenue, obeying traffic signals and maintaining safe distances." &
  "A car suddenly speeds through red lights, narrowly missing pedestrians and causing chaos." \\
 &
   &
  "Cars are moving at a consistent speed in their respective lanes, maintaining proper following distances." &
  "A vehicle hydroplanes, loses control, and collides with the median barrier." \\
 &
   &
  "Neighbors jog and walk their dogs while cars drive by steadily, without any aggressive behavior." &
  "A drunk driver swerves erratically, posing a serious threat to everyone on the road." \\ \hline
\end{tabularx}
\caption{Examples of prior knowledge generated by the LLM for different scenarios in the UCF-Crime dataset.}
\label{tab:ucf-know}
\end{table*}

\end{document}